\documentclass[10pt,twocolumn,letterpaper]{article}

\usepackage{wacv}
\usepackage{times}
\usepackage{epsfig}
\usepackage{graphicx}
\usepackage{amsmath}
\usepackage{amssymb}
\usepackage{algorithm}
\usepackage{algpseudocode}
\usepackage{enumitem}
\usepackage{xcolor}
\usepackage{subfigure}
\usepackage{multirow}
\usepackage{multicol}
\usepackage{colortbl}

\newcommand{\formattedparagraph}[1]{\noindent \textbf{#1}}

\def\wacvPaperID{671} 

\wacvfinalcopy 

\ifwacvfinal
\def\assignedStartPage{1} 
\fi

\ifwacvfinal
\usepackage[breaklinks=true,bookmarks=false]{hyperref}
\else
\usepackage[pagebackref=true,breaklinks=true,colorlinks,bookmarks=false]{hyperref}
\fi

\ifwacvfinal
\else
\fi

\begin{document}

\title{Neural Radiance Fields Approach to Deep Multi-View Photometric Stereo}

\author{\quad Berk Kaya$^1$\quad Suryansh Kumar$^1$ \quad Francesco Sarno$^1$\quad Vittorio Ferrari$^{2}$\quad Luc Van Gool$^{1, 3}$\\
Computer Vision Lab, ETH Z\"urich${^1}$, Google Research$^2$, KU Leuven$^3$
}

\maketitle

\begin{abstract}
We present a modern solution to the multi-view photometric stereo problem (MVPS). Our work suitably exploits the image formation model in a MVPS experimental setup to recover the dense 3D reconstruction of an object from images. We procure the surface orientation using a photometric stereo (PS) image formation model and blend it with a multi-view neural radiance field representation to recover the object's surface geometry. Contrary to the previous multi-staged framework to MVPS, where the position, iso-depth contours, or orientation measurements are estimated independently and then fused later, our method is simple to implement and realize. Our method performs neural rendering of multi-view images while utilizing surface normals estimated by a deep photometric stereo network. We render the MVPS images by considering the object's surface normals for each 3D sample point along the viewing direction rather than explicitly using the density gradient in the volume space via 3D occupancy information. We optimize the proposed neural radiance field representation for the MVPS setup efficiently using a fully connected deep network to recover the 3D geometry of an object. Extensive evaluation on the DiLiGenT-MV benchmark dataset shows that our method performs better than the approaches that perform only PS or only multi-view stereo (MVS) and provides comparable results against the state-of-the-art multi-stage fusion methods.
\end{abstract}


\section{Introduction}\label{sec:intro}

\begin{figure}
    \begin{center}
      \includegraphics[width=0.45\textwidth]{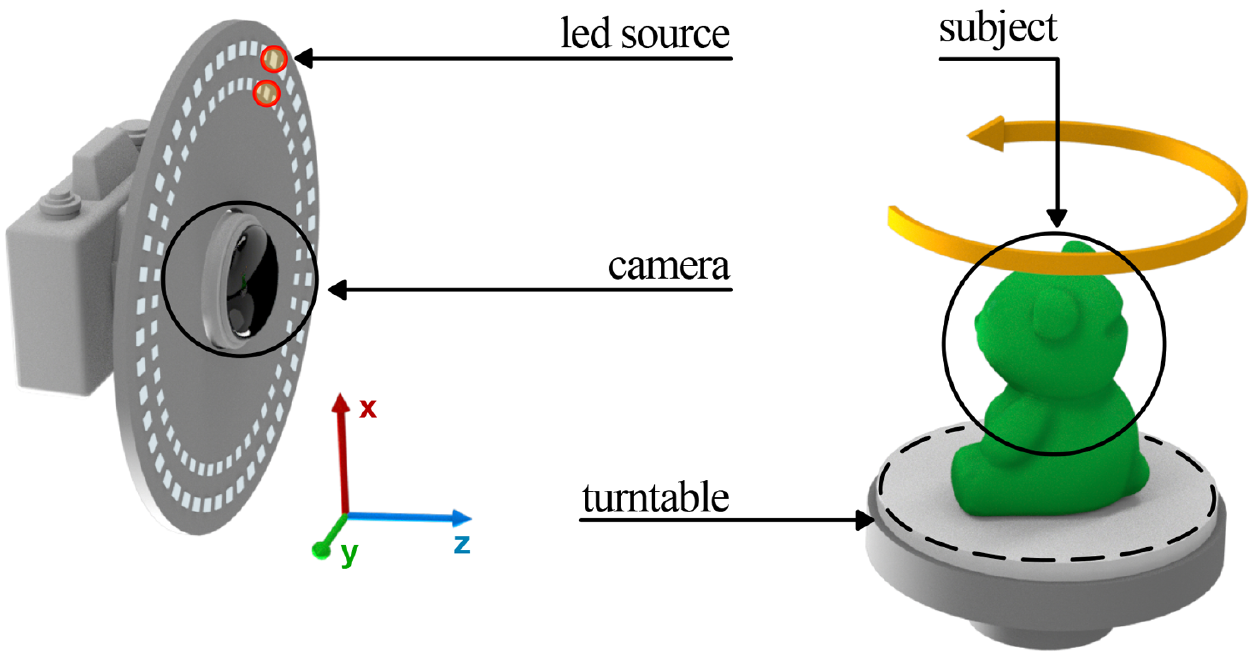}
    \end{center}
    \caption{\small General setup for multi-view photometric stereo. The object is placed on a base with fixed rotation. A camera is placed at the center of the disk to capture images. LEDs are placed in a concentric ring for controlled light setup \cite{li2020multi}.}\label{fig:MVPS_setup}
    \vspace{-0.25cm}
\end{figure}

Estimating the 3D geometry of an object from images has been a fundamental research problem in computer vision for several decades \cite{blake1987visual, ikeuchi1983constructing, ikeuchi1987determining}. A reliable solution has many real-world applications such as shape analysis, shape manipulation, molding, image rendering, forensics, \etc. Even though there exist several ways to solve this problem, multi-view photometric stereo is one of the popular setups to recover the 3D geometry of the object (see Fig.\ref{fig:MVPS_setup}).

In the past, many active and passive 3D reconstruction approaches or pipelines were proposed to solve 3D reconstruction of objects \cite{moons20093d, szeliski2010computer, furukawa2015multi, zollhofer2018state, menini2021real, kumar2018scalable, kumar2019jumping}. However, when it comes to the accuracy of recovered 3D shapes for its use in scientific and engineering purposes (metrology),  methods that use only MVS or PS  suffer \cite{mildenhall2020nerf, schonberger2016structure, furukawa2009accurate, galliani2015massively}. As a result, a mixed experimental setup such as multi-view photometric stereo (MVPS) is generally employed \cite{li2020multi}. In such a setup, complementary modalities are used to obtain better surface measurements, which are otherwise unavailable from an individual sensor or method. Accordingly, similar fusion-based strategies gained popularity for surface estimation \cite{mostafa1999integrating, zollhofer2018state, lange1999advances, nehab2005efficiently, chatterjee2015photometric}. One may also prefer to use two or more active sensors to receive the surface data estimates for fusion. Nevertheless, this paper focuses on the MVPS setup, where the subject is placed on a rotating base and for each rotation multiple images are captured using one LED light source at a time. The major motivation for such an approach is that the active range scanning strategies used for object's 3D acquisition, such as structured light \cite{geng2011structured, zhang2002rapid, zhang2003spacetime}, 3D laser scanners \cite{davis2003spacetime}, RGB-D sensors \cite{zollhofer2018state} are either complex to calibrate or provide noisy measurements or both. Further, these measuring techniques generally provide incomplete range data with outliers that require serious efforts for refinement.

Among the passive methods \cite{schoenberger2016sfm, wu2011visualsfm, furukawa2009accurate, triggs1999bundle, chatterjee2015photometric, kumar2017monocular, kumar2019superpixel, kumar2019dense}, Multi-View Stereo (MVS) is a key approach in the automated acquisition of 3D objects from a set of images taken from different viewpoints \cite{furukawa2009accurate}. Under the assumption of scene rigidity (rigid Lambertian textured surfaces), traditional MVS uses feature correspondences across images to reconstruct dense object geometry \cite{furukawa2015multi}. Recently, neural view synthesis methods have shown great potential for this task. However, the quality of their 3D reconstruction is far from satisfactory and several follow-ups are trying to address them for different scenarios \cite{mildenhall2020nerf, yu2020pixelnerf, yu2021plenoctrees, yariv2020multiview, chen2021mvsnerf}.





An alternative approach to recover surface details is photometric stereo (PS), a.k.a shape from shading \cite{woodham1980photometric, belhumeur1999bas}. PS methods estimate the object's surface normals given multiple images illuminated by different light sources, but all captured from the same camera viewpoint \cite{woodham1980photometric}. It is excellent at recovering surface normals independent of depth estimates and works well for untextured objects, non-Lambertian surfaces with fine details \cite{hertzmann2005example, wu2010robust, ikehata2014photometricisotropic, ikehata2018cnn, chandraker2005reflections, chen2019self}. To estimate the overall shape of the object, we can directly integrate the high-resolution normal map. However, the absence of constraints between surface elements and the lack of global geometric constraints fails to determine each of those surface components' relative positions correctly.  Further, most photometric stereo methods assume isotropic material objects and may fail to handle objects with anisotropic material like a piece of wood \cite{ikehata2018cnn, taniai2018neural, shi2013bi, ikehata2014photometricisotropic}.

Hence, MVS and PS complementary behavior in surface reconstruction from images helps us efficiently recover object shape. Using MVPS images, we utilize a CNN-based PS method to estimate surface normals using multiple light sources from a fixed viewpoint. For each view, multiple light sources are used to estimate the surface normal. We blend those surface normal estimates with a neural radiance field representation of the scene to recover the 3D reconstruction of the object.
%
Existing state-of-the-art methods to this problem generally apply a sequence of steps \cite{nehab2005efficiently, li2020multi, ren2021complex}
\textit{(i)} procure the 3D position measurement using multi-view images or 3D scanner \textit{(ii)} estimate surface orientation or iso-depth contours using photometric stereo methods, and
\textit{(iii)} fuse the surface orientation and 3D position estimates to recover 3D geometry using appropriate mathematical optimization.
Now, several fusion strategies exist that can combine these alternate sources of information for better 3D shape reconstruction. \cite{nehab2005efficiently, mostafa1999integrating, bylow2019combining, park2016robust, li2020multi, ren2021complex}. Of course, the precise steps taken by such approaches can provide better results, yet they rely heavily on explicit mathematical modeling \cite{bylow2019combining, nehab2005efficiently, chatterjee2015photometric, yu2013shading, hernandez2008multiview, park2016robust, li2020multi}, and \textbf{complex multi-staged network design} \cite{ren2021complex} which are complicated to execute.
In contrast, our work provides a \textbf{simple} and general method that can suitably introduce the surface details coming from PS to neural radiance field representation of the scene for favorable performance gain. 


Our work is inspired by the idea of local region-based techniques for volumetric illumination, which can render more realistic images \cite{levoy1988display, jonsson2014survey}. The ray-traced volume rendering approximates the surface normals using the gradient of the density along each $(x, y, z)$ direction in volumetric space \cite{pfister2005hardware, jonsson2014survey}. However, to use such a notion for our problem setup, we must know the occupancy of the point in the volume space.  To keep it simple, we utilize the surface normal from PS as the gradient density information for each sample point along the ray to reconstruct MVPS images as close as possible and recover 3D geometry. While one could recover depth from surface normals and then infer the occupancy of the volume density, we know that normal integration may lead to inaccurate depth estimates, hence incorrect occupancy information. So, we adhere to the proposed idea of using local gradients and show the validity of our method via experimental evaluations.

Our approach first estimates the surface normal for all the views using a deep photometric stereo network which is trained independently in a supervised setting. We use sample spatial location $\mathbf{x}_i$, the viewing direction $\mathbf{d}$, and the object's surface normal for our MVPS representation. Surface normals for each 3D sample point along all known view directions are employed. For 3D reconstruction, our method optimizes a multi-layer perceptron that regresses from the position, view direction, and surface normal to single volume density and view-consistent RGB color. We use Fourier feature encoding on both positional and surface orientation data before passing them to the network \cite{rahaman2019spectral}.
Experimental results show that our method provides comparable or better results than previously proposed complex multi-view photometric stereo methods \cite{bylow2019combining, nehab2005efficiently, chatterjee2015photometric, yu2013shading, hernandez2008multiview, mostafa1999integrating}, stand-alone state-of-the-art multi-view method \cite{wang2020patchmatchnet}, and view synthesis approach \cite{mildenhall2020nerf}. 
We evaluate the results on the DiLiGenT MVPS benchmark dataset \cite{li2020multi, shi2019tpami}\footnote{This dataset is publicly available for research purpose at: https://sites.google.com/site/photometricstereodata/mv}. We make the following contributions:
\begin{itemize}[leftmargin=*,topsep=0pt, noitemsep]
    \item While previous multi-stage fusion methods to MVPS are very complex, we propose a much simpler continuous volumetric rendering approach that uses the local density gradient effects in MVPS image formation.
    \item Our work takes an {opportunistic approach} that exploits the complementary source of information in MVPS setup via two sets of representation \ie, volumetric and surface.
    \item Despite being much simpler than fusion methods, our approach achieves better or comparable results than the state-of-the-art \cite{park2016robust, li2020multi}, as well as stand-alone methods such as multi-view \cite{wang2020patchmatchnet, yariv2020multiview}, photometric stereo \cite{ikehata2018cnn}, and continuous volumetric rendering \cite{martin2020nerf} methods.
\end{itemize}

\section{Related Work}\label{sec:related_works}
Here, we review important MVS, PS, view-synthesis and fusion based methods related to our work.

\formattedparagraph{1. Multi-View Stereo (MVS) Methods.}
It aims at reconstructing a plausible 3D geometry of the object from a set of images alone \cite{furukawa2015multi}. The working principle of MVS generally pivots around improving the dense image feature correspondence (local similarity) and camera parameters across images for triangulation \cite{furukawa2015multi, kutulakos2000theory, furukawa2009accurate, bleyer2011patchmatch, kumar2017monocular, kumar2019superpixel}. Recent developments in machine learning have led to the renovation of traditional MVS methods via deep-learning frameworks. 

Roughly, it can be divided into four to five categories \cite{furukawa2015multi, gu2020cascade}. (\textit{i}) Volumetric methods require bounding box knowledge containing the subject. These methods estimate the relation between each voxel and the surface element, and their accuracy is greatly affected by voxel grid resolution \cite{ji2017surfacenet, kar2017learning, seitz1999photorealistic, faugeras2002variational, vogiatzis2005multi}. (\textit{ii}) Patch-based approach utilize the Barnes \etal \cite{barnes2009patchmatch} randomized correspondence idea in the scene space. Generally, these methods generate random 3D planes in the scene space and refine the depth representation and normal fields based on photo-consistency measures \cite{bleyer2011patchmatch, galliani2015massively, furukawa2009accurate, locher2016progressive}. (\textit{iii}) Depth map reconstruction-based methods use a reference image with different source images under calibrated settings to improve the overall depth estimation \cite{goesele2006multi, strecha2006combined, campbell2008using, schonberger2016pixelwise, galliani2015massively, xu2019multi}. (\textit{iv}) Point cloud-based methods operate on the 3D points and process the initial sparse point sets to densify the results \cite{furukawa2009accurate, lhuillier2005quasi}. (\textit{v}) Distributed structure-from-motion methods utilize the notion of motion averaging to improve large-scale 3D reconstruction \cite{chatterjee2013efficient, Zhang2017ICCV, zhu2018very}. Recently, deep neural networks have been widely adopted for MVS, which provide better performance than traditional MVS methods \cite{gu2020cascade}. Earlier work in this area uses CNN's for two-view \cite{zbontar2016stereo} and multi-view stereo \cite{hartmann2017learned}. Lately, the learning-based MVS rely on the construction of 3D cost volume and use the deep neural networks for regularization and depth regression \cite{chen2019point, hou2019multi, xue2019mvscrf, im2018dpsnet, luo2019p, yao2018mvsnet, xu2020learning}. As most of these approaches utilize 3D CNN for cost volume regularization ---which in general is computationally expensive, the majority of the recent work is motivated to meet the computational requirement with it. Few methods attempt to address it by down-sampling the input \cite{yao2018mvsnet, xu2020learning}. Other attempts to improve the computational requirements uses sequential processing of cost volume \cite{yao2019recurrent}, cascade of 3D cost volumes \cite{cheng2020deep, gu2020cascade, xu2020pvsnet, wang2020patchmatchnet}, small cost volume with point-based refinement \cite{chen2019point}, sparse cost volume with RGB and 2D CNN to densify the result \cite{yu2020fast}, learning-based patch-wise matching \cite{luo2019p, wang2020patchmatchnet} with RGB guided depth map super-resolution \cite{wang2020patchmatchnet}.

\formattedparagraph{2. Photometric Stereo (PS) Methods.}
It is an alternative approach to infer the surface geometry from light-varying images. Those images are captured from a camera placed at a fixed viewpoint \cite{woodham1980photometric}. Although PS requires a controlled experimental setup, it prevails in recovering surface details independent of the depth knowledge. Similar to MVS, advancements in deep learning have led to the development of deep PS methods \cite{ikehata2018cnn, chen2020deep, taniai2018neural, chen2019self}. The primary motive of the deep PS method is to let the neural network learn the complicated surface reflectance from data, which otherwise is a challenging modeling task for surfaces with unknown BRDF's \cite{taniai2018neural, kaya2020uncalibrated}.

Woodham's introduced PS for Lambertian surfaces \cite{woodham1980photometric}. Since then, the majority of the work in PS pivot around improving it for non-Lambertian surfaces \cite{georghiades2003incorporating, chung2008efficient, goldman2009shape, ikehata2014photometricisotropic, wu2010robust}. In general, traditional PS can be categorized into four groups. (\textit{i}) Robust methods recover surface normals by assuming the simple diffuse reflectance property of the object. It treats non-Lambertian reflectance components to be local and sparse, hence treated as outliers. Accordingly, rank minimization \cite{wu2010robust}, expectation-maximization \cite{wu2009photometric}, sparse Bayesian regression \cite{ikehata2014photometric}, RANSAC \cite{mukaigawa2007analysis}, and other outlier filtering mechanisms \cite{miyazaki2010median} are used to estimate surface normals. Yet, these methods cannot handle material with soft and broad specularity \cite{shi2019tpami}. (\textit{ii}) Analytic BRDF modeling methods explicitly model the object's specularity as a sum of diffused and specular components \cite{georghiades2003incorporating, chung2008efficient, goldman2009shape}. Still, it applies only to a limited class of objects due to the hand-crafted modeling strategy. (\textit{iii}) General reflectance modeling-based methods utilize the objects isotropic \cite{shi2013bi, ikehata2014photometric} and anisotropic \cite{holroyd2008photometric} properties to estimate unknown BRDF rather than explicitly modeling it. (\textit{iv}) the example-based method uses known object's material knowledge for reflectance modeling \cite{silver1980determining}. While some work also assumes that the object's material can be expressed using a small number of basis material \cite{hertzmann2005example}, others take advantage of virtual sphere rendered with different materials without using any physical reference object \cite{hui2016shape}.
Recent PS research is aimed at learning the complex surface reflectance using data via deep neural networks. Early deep PS methods aimed at learning the mapping between surface normals and reflectance measurements in a calibrated setting \ie, light sources direction is given at train and test time \cite{santo2017deep, taniai2018neural, ikehata2018cnn, chen2018ps, zheng2019spline, li2019learning}. Recently, uncalibrated deep PS methods have been proposed to overcome the challenges associated with light sources calibration \cite{chen2020deep, chen2019self, kaya2020uncalibrated} and resolve GBR ambiguity in photometric stereo \cite{chen2020learned}.

\formattedparagraph{3. View Synthesis for 3D reconstruction.} In recent years, view synthesis methods, in particular, the Neural Radiance Fields (NeRF) method for scene representation, have proposed an interesting idea to recover the 3D geometry of the scene using multi-view images \cite{mildenhall2020nerf}. NeRF has generated a new wave of interest in 3D computer vision and has led to several follow-ups in 3D data acquisition from images \cite{yu2020pixelnerf, martin2020nerf, neff2021donerf, pumarola2020d, srinivasan2020nerv, yu2021plenoctrees, rematasICML21, chen2021mvsnerf}. NeRF uses a fully-connected deep neural network to represent the scene geometry and radiance information implicitly. It renders photo-realistic views by implicitly encoding surface volume density and color via a multi-layer perceptron (MLP). Once MLP is trained, the 3D shape can be recovered by using the estimated volume density. One can model the implicit object's surface normals and radiance field in a unified way but may need volume occupancy information. Further, to couple the surface normals term with the surface 3D point coming from the same measurement source may not provide significant gain \cite{yariv2020multiview}, especially when dealing with PS images, where shading plays a critical role.

\formattedparagraph{4. Fusion based Methods.}
Although the developments in RGB-D and other portable active sensors have led to success with volumetric methods on shading-based refinement \cite{bylow2019combining, maier2017intrinsic3d, zollhofer2015shading}, our work relies on image data for this problem. Much of the existing work that utilizes the images for complementary measurements uses explicit mathematical modeling for precise 3D geometry \cite{hernandez2008multiview, park2016robust, li2020multi, logothetis2019differential, liang2020better, ren2021complex}. \textit{On the contrary}, our work utilizes photometric stereo to estimate surface normals and then blends it with the shape volume density in neural radiance field representation to synthesize multi-view images and recover shapes. Table (\ref{tab:fusion_methods}) summarizes some of the recent and early developments in the area of multi-view photometric stereo.  

\section{Proposed Approach}\label{sec:our_approach}
We denote $\mathcal{I}^{v} = \{I_1^{v},.., I_{N_p}^{v}\}$ as the set of $N_p$ PS images for a given view $v \in [1, N_m]$. Our method assumes a calibrated setting for MVPS, \ie, all the light source directions and camera calibrations are known. The proposed approach considers the notable image formation models used in computer vision and computer graphics \ie, photometric stereo image formation model \cite{woodham1980photometric} and the rendering equation \cite{kajiya1986rendering}. These imaging models has its advantage depending on the experimental setup. Since we are solving the well-known MVPS problem, we can exploit the benefit of both photometric and multi-view stereo setup. For our work, we assume that the subject under study is a solid texture-less object, which is often studied in photometric stereo research \cite{li2020multi}. Next, we describe the deep PS network, followed by our neural radiance field modeling for a multi-view photometric stereo setup.

\formattedparagraph{Deep Photometric Stereo.} For simplicity, let's assume a single view case for an object with reflective surface, whose appearance can be encoded by a  bidirectional reflectance distribution function (BRDF) $\Phi_s$ with surface normals $\mathbf{N} \in \mathbb{R}^{ 3 \times p}$ \cite{taniai2018neural, kaya2020uncalibrated}. Here, $p$ symbolizes the total number of  pixels. When object's surface is illuminated by a point light source positioned in the direction $l_i \in \mathbb{R}^{3 \times 1}$, then the image $I_i \in \mathbb{R}^{c \times r}$ captured by a camera in the view direction $v \in \mathbb{R}^{3 \times 1}$ can be modeled as
\begin{equation}
    \centering
    \begin{aligned}
    I_{i}^{v} = e_{i} \cdot \Phi_s(\mathbf{N}, l_i,  v) \cdot \max(\mathbf{N}^{T}l_i, 0)  + \epsilon_i  
    \end{aligned}\label{eq:image_formation_model_ps}
\end{equation}
where, $e_i \in \mathbb{R}_+$ denotes the intensity, $(c, r)$ symbolizes the columns and rows of an image, $\epsilon_i$ is an additive error and $\max(\mathbf{N}^{T}l_i, 0)$ accounts for the attached shadows. Eq.\eqref{eq:image_formation_model_ps} image formation model for photometric stereo has led to outstanding developments in recovering fine details of the surface \cite{shi2016benchmark}. Yet, modeling unknown reflectance properties of different objects remains a fundamental challenge. Consequently, we utilize deep neural networks to learn the complicated BRDF's from input data. We leverage the observation map based CNN model to estimate surface normal under a calibrated setting \cite{ikehata2018cnn}. Unlike other supervised methods, it has rotational invariance property for isotropic material, handles unstructured images and lights well, and above all provides best performance known to us with acceptable inference time.

\begin{table}[t]
\centering
\resizebox{\columnwidth}{!}
{\begin{tabular}{|c|c|c|c|}
\hline
\rowcolor[gray]{0.85}
\textbf{Method} &  \textbf{Base data}  &  \textbf{Shape representation}  & \textbf{Optimization} \\
\hline
Mostafa \etal \cite{mostafa1999integrating} & Laser, PS images  & Neural Network  & Back-Propagation + EKF \cite{wan2001dual}  \\
\hline
Nehab \etal \cite{nehab2005efficiently} & Scanner \cite{davis2003spacetime}, PS images  &  Mesh &  Mesh (Linear)\\
\hline
Hernandez \etal \cite{hernandez2008multiview} &  MVPS images & Mesh & Mesh (Coupled)\\ 
\hline
Park \etal \cite{park2016robust} & MVPS images & Mesh + 2D displacement & Parameterized 2D (Sparse Linear)  \\ 
\hline
Li \etal \cite{li2020multi} & MVPS images  & Depth Contour + 3D points  & Poisson surface \cite{kazhdan2006poisson} + \cite{nehab2005efficiently}  \\
\hline
Logothetis \etal \cite{logothetis2019differential} & MVPS images and SDF  & Parameterized SDF & Variational  Approach\\
\hline
Ours &  MVPS images & Multi-layer Perceptron  & Adam \cite{kingma2014adam} \\
\hline
\end{tabular}}
\caption{\small Previous work on passive approach to 3D shape recovery using orientation and range estimates. Despite Mostafa \etal \cite{mostafa1999integrating} and Nehab \etal  \cite{nehab2005efficiently} use active modality, we included them for completeness. }
\label{tab:fusion_methods}
\end{table}

\textit{$\triangleright$ {Observation map}:} For each pixel, this map contains the normalized observed intensity values due to all the light sources. In a general PS setup, the light sources are located in a concentric way. Thus, a one-to-one mapping between the light source position $(l_x, l_y, l_z) \in \mathbb{R}^3$ and corresponding x-y coordinate projection $(l_x, l_y) \in \mathbb{R}^{2}$ is possible. Note that $l_x^2 + l_y^2 + l_z^2 = 1 ~\forall ~{l}_i$ \ie, the unit vector in the direction of source. We construct the observation map $\Omega_j \in \mathbb{R}^{w \times w}$ for each pixel $j$ using its intensity value across all the $N_p$ images as outlined in \textbf{Algorithm 1}.
Here, $w$ is the size of the observation map and the function $\zeta: \mathbb{R} \mapsto \mathbb{Z}_{0+}$. The scalar $\eta_j$ for $j^{th}$ pixel is $\eta_{j} = \max(e_1/I_{1}^{v}(j),..,e_{N_p}/I_{N_p}^{v}(j))$ is the normalizing constant. Since the projected source vectors can take values from $[-1, 1]$, they are scaled appropriately to get positive integer values.

\setlength{\intextsep}{8pt}%
\setlength{\columnsep}{10pt}%

\begin{algorithm}
\footnotesize \caption{\footnotesize Observation Map Construction}
\begin{algorithmic}
\State $\textbf{for} ~~j \gets \{1,., p\}$ 
\State ~~$\textbf{for} ~~i \gets \{1,., N_p\}$
\State ~~~~$~\Omega_{j}\Big(\zeta\big(w\cdot \frac{(l_x^{i} + 1)}{2} \big), \zeta\big(w\cdot \frac{(l_y^{i} + 1)}{2} \big) \Big) = \eta_j \frac{I_i^{v}(j)}{e_i}$
\State ~~$\textbf{end}$
\State  $\textbf{end}$
\end{algorithmic}
\end{algorithm}

\begin{figure*}[t]
    \centering
    \includegraphics[{width=1.0\linewidth}]{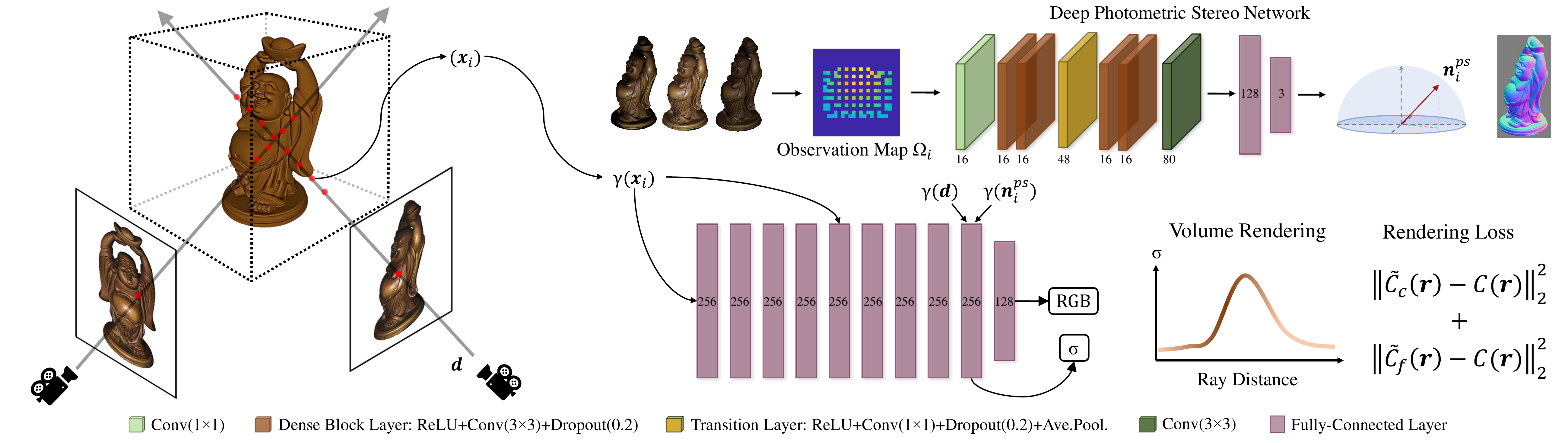}
    \caption{ \small \textbf{Overview}: The deep photometric stereo network predicts surface normals of the object from each viewpoint using PS images. We model multi-view neural radiance fields by introducing gradient knowledge from PS network output in the density space for solving MVPS. Our work takes a much simpler approach than existing state-of-the-art multi-staged MVPS methods showing comparable accuracy.}
    \label{fig:pipeline}
\end{figure*}

\textit{$\triangleright$ {PS Neural Network Architecture Details}:} Inspired by the DenseNet design \cite{huang2017densely}, the PS network first performs a convolution with 16 output channels on the input observation maps. The other part of the network consists of two dense blocks, a transition layer, and two dense layers, which is then followed by a normalization layer to recover surface normals as output. The dense block is composed of one ReLU layer, one $3 \times 3$ convolutional layer and a 0.2 dropout layer. The transition layer is composed of a ReLU layer, $1 \times 1$ convolutional layer, 0.2 dropout layer, and an average pooling layer, which is placed in between the two dense blocks to modify the feature map size. We train PS network end-to-end separately in a supervised setting. The $l_2$ (MSE) loss between estimated and the ground-truth normals is minimized using Adam optimizer \cite{kingma2014adam} (see Fig. \ref{fig:pipeline}). 

\formattedparagraph{Neural Radiance Field Representation for Multi-View Photometric Stereo.} Recently, volume rendering techniques for view synthesis, in particular NeRF \cite{mildenhall2020nerf} has shown a great potential for 3D data-acquisition from multi-view images. It represent a continuous scene as a 5D vector-valued function \ie, $\mathbf{x} = (x, y, z)$ for each 3D location and $(\theta, \phi)$ for every 2D viewing direction. Given multi-view images with known camera pose, NeRF approximate the assumed continuous 5D scene representation with a MLP that maps the $(\mathbf{x}, \theta, \phi)$ to RGB color $\mathbf{c}$ and volume density $\mathbf{\sigma} \in \mathbb{R}_{+}$. Using the classical volume rendering work \cite{kajiya1984ray}, it models the expected color $C(\mathbf{r})$ of the camera ray $r(t) = \mathbf{o}+t\mathbf{d}$ with near and far bound $t_n, t_f$ as:

\begin{equation}
    \centering 
    C(\mathbf{r}) = \int_{t_n}^{t_f} T(t) \sigma \big(\mathbf{r}(t)\big) \mathbf{c}\big( \mathbf{r}(t), \mathbf{d}\big)dt,
    \label{eq:nerf_cont_eq}
\end{equation}

\begin{equation}
    \centering 
    ~\text{where} ~T(t) = \exp\Big(- \int_{t_n}^{t} \sigma \big(\mathbf{r}(s)\big)ds\Big) .
    \label{eq:nerf_cont_eq_transmittance}
\end{equation}
Here, $\mathbf{d}$ is the unit viewing direction. $T(t)$ is the accumulated transmittance along ray from $t_n$ to $t$ which caters the notion that how much light is blocked earlier along the ray. Given $N$ samples along the ray, the continuous integral in Eq:(\ref{eq:nerf_cont_eq}) is approximated using the quadrature rule \cite{max1995optical}:
\begin{equation}
    \centering
    \Tilde{C}(\mathbf{r}) \approx \sum_{i=1}^{N} T_i \alpha_i(\mathbf{x}_i) \mathbf{c}_i (\mathbf{x}_i, \mathbf{d}),
   \label{eq:color_approx}
\end{equation}
\begin{equation}
    \centering
    \text{where} ~\alpha_i (\mathbf{x}_i) = \Big(1-\exp\big(\sigma(\mathbf{x}_i)\delta_i\big)\Big),         
    \label{eq:color_approx_alpha}
\end{equation}
\begin{equation}
    \centering
     ~\text{and} ~T_i = \prod_{j=1}^{i-1}(1-\alpha_j).
        \label{eq:color_approx_T}
\end{equation}
Here, $\delta_i$ is the distance between the adjacent discrete samples and $\alpha_i$ encapsulate how much light is contributed by the ray $i$. By construction, Eq.\eqref{eq:color_approx} approximates the alpha composited color as a weighted combination of all sampled colors $\mathbf{c}_i$ along the ray. For more details we refer the readers to Mildenhall \etal work \cite{mildenhall2020nerf}. Now, lets have a closer look at the following rendering equation \cite{kajiya1986rendering, yariv2020multiview}:
\begin{equation}
    \centering
    \begin{aligned}
      L_o ( & \mathbf{x}_i, \omega_o)  =  L_e(\mathbf{x}_i, \omega_o)  \\
     & + \int_{\mathcal{S}} \Phi_r(\mathbf{x}_i, \mathbf{n}_i, \omega_j, \omega_o) L_j(\mathbf{x}_i, \omega_j)(\mathbf{n}_i \cdot \omega_j) d\omega_j .
    \end{aligned}\label{eq:uni_rendering_equation}
\end{equation}
The above equation suggests that the rendering of a surface element $\mathbf{x}_i$ depends on the emitted light in the scene and the bidirectional reflectance distribution function (BRDF) $\Phi_r$ describing reflectance and the color property of the surface accumulated over the half-sphere $\mathcal{S}$ centered at $\mathbf{n}_i$. The significance of including surface normal to $\Phi_{r}$ in the rendering equation is put forward by IDR work \cite{yariv2020multiview}. Here, $\omega_o, \omega_j$ is the outgoing light direction and negative direction of the incoming light, respectively. $\Phi_r$ accounts for proportion of light reflected from $\omega_j$ towards $\omega_o$ at $\mathbf{x}_i$. $L_o$ the radiance directed outward along $\omega_o$ from a particular position $\mathbf{x}_i$, and $L_e$ is the emitted radiance of the light. In general, the light does not always hit the surface orthogonally, and so the dot product between the $\mathbf{n}_i$ and $\omega_j$ attenuates the incoming light at $\mathbf{x}_i$. Hence, by restricting $\Phi_r$ and light radiance functions ($L_o, L_e$), which the radiance fields approximation can represent, we condition the color function to include the notion of density gradient for image rendering. Consequently, we use the normal estimated from PS to condition $\mathbf{c}_i$ in Eq.\eqref{eq:color_approx}. We rely on a deep photometric stereo network to estimate surface normals, overcoming BRDF modeling complications and providing us an excellent surface detail. Hence, our method has the inherent benefit over an entangled surface normal representation in image rendering \cite{yariv2020multiview}.  Accordingly, we modify the Eq.\eqref{eq:color_approx} as follows:
\begin{equation}
    \centering
    \begin{aligned}
    \Tilde{C}(\mathbf{r}) \approx \sum_{i=1}^{N} T_i \alpha_i(\mathbf{x}_i) \mathbf{c}_i (\mathbf{x}_i, \mathbf{n}_i^{ps}, \mathbf{d}) .
    \end{aligned}\label{eq:our_color_function}
\end{equation}
Further, adding image features in Eq.\eqref{eq:our_color_function} could be advantageous as in \cite{yu2020pixelnerf, chen2021mvsnerf}. However, MVPS setup generally deals with non-textured surfaces where using image features is not much of help. Still, an obvious advantage with the setup is that better surface details can be captured from shading. For simplicity, we use surface normals to condition volume rendering and refrain from relying on image features. So, our approach blends density gradient information into the continuous volume rendering formulation bypassing the explicit volume occupancy information. Concretely, we feed surface normals for each 3D sample point along the viewing direction to the neural rendering network.

\formattedparagraph{Optimization and Loss Function.}
Following neural radiance fields optimization strategy \cite{mildenhall2020nerf}, we encode each sampled position $\mathbf{x}_i$ along the ray, viewing direction $\mathbf{d}$, and photometric stereo surface normal $\mathbf{n}_i^{ps}$ using the Fourier features $\gamma(\mathbf{x}) = [\sin(\mathbf{x}), \cos(\mathbf{x}),.., \sin(2^{L-1}\mathbf{x}), \cos(2^{L-1}\mathbf{x})]$. We used $L=10$ for $\gamma(\mathbf{x}_i)$, $L=4$ for $\gamma(\mathbf{n}_i^{ps})$ and $L=4$ for $\gamma(\mathbf{d})$. For efficient estimation of continuous integral (Eq.\eqref{eq:our_color_function}) using quadrature rule, we use the stratified sampling approach to partition the near and far bound $[t_n, t_f]$ into $N$ evenly-spaced discrete samples  \cite{max1995optical}.
\begin{equation}
    \centering
    \begin{aligned}
     t_i \sim \mathcal{U} \Big[ t_n + \frac{i-1}{N}(t_f - t_n), t_n + \frac{i}{N}(t_f - t_n)\Big].
    \end{aligned}
\end{equation}

We employed MLPs (Multi-layer Perceptron) to optimize the following loss function.
\begin{equation}
    \centering
    \begin{aligned}
     \mathcal{L}_{mvps} = \sum_{\mathbf{r} \in \mathcal{B}} \|\Tilde{C}_c(\mathbf{r})- C(\mathbf{r})\|_2^2 + \|\Tilde{C}_f(\mathbf{r})- C(\mathbf{r})\|_2^2 .
    \end{aligned}\label{eq:our_loss}
\end{equation}
where, $\mathcal{B}$ denotes set of all rays in the batch. We used hierarchical volume sampling strategy to densely evaluate neural radiance field network for $N$ query point along each ray. To that end, we first sample $N_c$ points using the stratified sampling strategy and optimize the coarse network $\Tilde{C}_c(\mathbf{r})$ (Eq.\eqref{eq:our_color_function}). With the known output distribution of the coarse network, we sample $N_f$ points using inverse transform sampling to optimize the fine network $\Tilde{C}_f(\mathbf{r})$. In Eq.\eqref{eq:our_loss}, the variable $C(\mathbf{r})$ is the observed color for the pixel (see Fig.\ref{fig:pipeline}).

\section{Experiment and Ablations}\label{sec:experiments_ablation}
We conducted our experiments, statistical evaluations, and ablation study on the DiLiGenT-MV benchmark dataset \cite{li2020multi, shi2019tpami}. It consists of complex BRDF images taken from 20 viewpoints. For each viewpoint, 96 images are captured, each illuminated by a light source in a different known position (calibrated). The dataset includes 5 objects ({BEAR}, {BUDDHA}, {COW}, {POT2}, {READING}) with complex surface profiles and its images are captured under large illumination changes. For creating this dataset, the distance between the camera and object is set to 1500 mm \cite{li2020multi}.

\smallskip
\formattedparagraph{Implementation Details.}
\textit{(a) Deep-Photometric Stereo}: We train the deep photometric stereo network on the CyclesPS dataset \cite{ikehata2018cnn}. We used the Adam optimizer \cite{kingma2014adam} with a learning rate of $10^{-3}$ and trained for 10 epochs. A per-pixel observation map with a size of $32\times 32$ is used at train and test time. During testing, we applied the network on 96 PS images per subject from DiLiGenT-MV dataset.
\textit{(b) MLP Optimization for MVPS}: For each object in DiLiGenT-MV dataset, we optimize a dense fully-connected neural network that is composed of 8 fully-connected ReLU layers. Each layer is composed of 256 channels. In addition to the density output from the $8^{th}$ layer, 24 dimensional view-direction and surface normal Fourier features are then fed at the $9^{th}$ layer for rendering (see Fig.\ref{fig:pipeline}). We use $N_c = 64$ points for the coarse network and $N_f =128$ points for the fine network. We optimize the network parameters for 30 epochs with a batch size of 1024 rays and an initial learning rate of $10^{-4}$. This takes 7 hours per object on an NVIDIA GeForce RTX 2080 Ti with 11GB RAM.

\subsection{Baseline Comparison}\label{ss:baseline_comparison}
For comparison against the baseline methods, we classified them into two categories:
\textit{(a)} Standalone methods: It either uses MVS or PS set up to recover the 3D shapes of an object. \textit{(b)} Multi-Staged Fusion methods: It first recovers the sparse 3D point cloud of the object using multi-view images and surface orientation using fixed viewpoint PS images for each view. These spatial positions and orientations are then fused using a different method/pipeline to recover the 3D geometry.
To compare against the standalone methods, we pick the $4^{th}$ light source in DiLiGenT-MV setup. To recover the 3D shape using our method, we run our fine model by sampling points in $512^3$ volumetric space uniformly. The recovered density is queried using marching-cube algorithm \cite{lorensen1987marching} with $\sigma=10$. 
Table \ref{tab:numerical_comparison} compares the 3D reconstruction accuracy of our method to standalone and multi-stage fusion methods. We evaluate accuracy using the standard Chamfer-L1 distance metric between the recovered shape and the ground-truth shape after registration.\\
\noindent
\formattedparagraph{(a) Standalone Methods.} 
For standalone baselines comparison, we compare our method with the recent state-of-the-art in MVS, PS, and View-Synthesis.


\begin{table*}[t]
    \centering
    \resizebox{0.70\textwidth}{!}
    {
    \begin{tabular}{c|c|c|c|c||c|c|c}
    \hline
    \rowcolor[gray]{0.90}
        \multicolumn{1}{c|}{Method Type $\rightarrow$} & \multicolumn{4}{c||}{Standalone Methods  ($\downarrow$) }&  \multicolumn{2}{c|}{Multi-Stage Fusion Methods  ($\downarrow$) } & \multicolumn{1}{c}{} \\
        \hline
         Dataset & PM-Net \cite{wang2020patchmatchnet} & IDR \cite{yariv2020multiview} & NeRF \cite{mildenhall2020nerf} &  Ours  & R-MVPS \cite{park2016robust} & B-MVPS \cite{li2020multi} & Ours \\
        \hline
        BEAR  & 2.13 & 10.97 & 0.62  & 0.66 & 0.89 & 0.63 & 0.66 \\
        BUDDHA  & 0.72 & 5.18 & 0.99  & 1.00 & 0.64 & 0.40 & 1.00   \\
        COW & 1.68 & 6.45 & 0.92  &  \textbf{0.71} & 0.42  & 0.54 & 0.71 \\
        POT2  & 1.30 & 6.68 & 0.64  & \textbf{0.63} & 1.29 & 0.55 & 0.63 \\
        READING  & 1.64 & 7.05 & 1.22  & \textbf{0.82} & 0.98 & 0.85 & \textbf{0.82} \\
    \hline
    \end{tabular}
    }
    \caption{\small Quantitative 3D reconstruction accuracy comparison against the competing methods on DiLiGenT-MV benchmark. We used Chamfer-L1 metric to compute the accuracy. The statistics show that our method is better and comparable to the stand-alone methods and Multi-Stage fusion methods respectively. \textbf{Generally, Multi-Stage fusion methods are heuristic in nature and require careful execution at different stages.} On the contrary, our method is much simpler to realize and implement. Note: Even though we tried our best to make sure that IDR codes are compiled the way it's on their paper code, we are surprised by IDR results and we are still investigating it. The above result is the best we achieved out of its publicly available code after several trials.}
        \label{tab:numerical_comparison}
\end{table*}


\noindent
\textit{(1) PatchMatch Network} \cite{wang2020patchmatchnet}:
This method has recently shown state-of-the-art performance in MVS. PatchMatch-Net proposed an end-to-end trainable network that has fast inference time and works well even for high-resolution images. It is trained on the DTU dataset \cite{aanaes2016large}.
We empirically observed that using two source frames per reference view on the DiLiGenT-MV dataset results in more accurate depth estimation. After getting the depth map from the network for each view, we transform the results to a point cloud by back-projecting the depth values to 3D space.

\noindent
\textit{(2) IDR} \cite{yariv2020multiview}:
This recently published method proposes a differentiable rendering formulation that is capable of implicitly modeling variety of lights and surface materials. IDR demonstrated state-of-the-art results on the DTU dataset and showed impressive 3D reconstructions from MVS images. For comparison, we train the model on DiLiGenT-MV for 2000 epochs. During training and testing, we maintained the default settings introduced by \cite{yariv2020multiview}.

\noindent
\textit{(3) NeRF} \cite{martin2020nerf}:
Even though it is developed for novel view synthesis, recently NeRF has been widely used as a baseline for multi-view 3D reconstruction \cite{yu2020pixelnerf}. By sampling the volume density $\sigma$, it is possible to recover 3D geometry. We use an initial learning rate of $10^{-4}$ and batch size of $1024$ rays. We run the optimization for 30 epochs and threshold the density values ($512^3$) at 10 to recover the 3D geometry. 

\begin{table}[t]
 \scriptsize
	\centering
		\begin{tabular}{c|c|c|c|c|c}
			\hline
			\rowcolor[gray]{0.85}
			Method & BEAR & BUDDHA & COW  & POT2 & READING \\ \hline
			CNN-PS \cite{ikehata2018cnn} & 0.78 & 0.83 & 0.87 & 0.86 & 0.89 \\ \hline
			Ours & \textbf{0.09} & \textbf{0.11} & \textbf{0.11} & \textbf{0.07} & \textbf{0.06} \\ \hline
		\end{tabular}
		\caption{\small  Multi-view depth error comparison against CNN-PS \cite{ikehata2018cnn}. For comparison, we integrate the surface normals from CNN-PS to recover its depth which is scaled appropriately to compute the $l_1$ depth accuracy. For ours, we projected the recovered shape reconstruction to the estimated depth and corresponding depth accuracy.}\label{tab:cnnps_comparison}
\end{table}

\noindent
\textit{(4) CNN-PS} \cite{ikehata2018cnn}: This method proposes a dense convolution neural network to learn the mapping between PS images and surface normals directly. It can handle non-convex surfaces and complicated BRDFs. We integrate the obtained surface normals using Horn and Brooks' method \cite{horn1986variational} to get the depth map, and scale it to [-1, 1].  Next, we also project our 3D shape to the different cameras and recover the depth on a similar scale for statistical comparison. Table \ref{tab:cnnps_comparison} shows our performance comparison against CNN-PS.



\begin{table*}[t]
    \small
    \centering
    \resizebox{0.70 \textwidth}{!}
    {
    \begin{tabular}{c|c|c||c|c||c|c||c|c||c|c}
    \hline
    \rowcolor[gray]{0.90}
        \multicolumn{1}{c|}{Dataset $\rightarrow$} & \multicolumn{2}{c||}{BEAR} &  \multicolumn{2}{c||}{BUDDHA} & \multicolumn{2}{c||}{COW} & \multicolumn{2}{c||}{POT2} & \multicolumn{2}{c}{READING}  \\
        \hline
         Method & PSNR $\uparrow$  & LPIPS $\downarrow$ & PSNR $\uparrow$ &  LPIPS $\downarrow$  & PSNR $\uparrow$  & LPIPS $\downarrow$  & PSNR $\uparrow$ & LPIPS $\downarrow$ & PSNR $\uparrow$ & LPIPS $\downarrow$ \\
        \hline
        IDR \cite{yariv2020multiview}  & 4.43 & 0.2370 & 9.87  & 0.2261 & 9.15 & 0.1571 & 7.71 & 0.1662 & 6.66 & 0.1815\\
        NeRF \cite{mildenhall2020nerf} & 29.97 & 0.0235 & 29.00  & 0.0455 & 30.80 & 0.0192 & 28.88 & 0.0269 & 28.12 & 0.0346\\
        Ours & \textbf{37.16} & \textbf{0.0122} & \textbf{33.59}  & \textbf{0.0162} & \textbf{34.49}  & \textbf{0.0134} & \textbf{30.47} & \textbf{0.0258} & \textbf{30.46} & \textbf{0.0311} \\
    \hline
    \end{tabular}
    }
    \caption{\small Quantitative image rendering quality comparison on DiLiGenT-MV benchmark. Our method can render much better images as it utilizes the surface normal information acquired with PS.  }
    \label{tab:psnr_lpips}
\end{table*}

\formattedparagraph{(b) Multi-Stage Fusion Methods.}
Fusion approaches to MVPS usually comprise several steps that are heuristic in nature, and proper care must be taken to execute all the steps well. For evaluation, we compared against the two well-known baselines in MVPS (see Table \ref{tab:numerical_comparison}).

\noindent \textit{(1) Robust MVPS} \cite{park2016robust}.
It employs a series of different algorithms to solve MVPS. It first uses multi-view images to recover the coarse 3D mesh of the object using structure from motion. Next, this mesh is projected to 2D planar space for parameterization to estimate multi-view consistent surface normals using photometric stereo setup. Finally, the estimated surface normals are used for mesh refinement to recover the fine-detailed geometry of the object. \\
\noindent
\textit{(2) Benchmark MVPS} \cite{li2020multi}. This method takes a set of steps to estimate a fine-detailed 3D reconstruction of an object from MVPS images. It first estimates iso-depth contours from the PS images \cite{alldrin2007toward} and then uses a structure-from-motion algorithm to recover a sparse point cloud from the MVS images. Later the depth of these 3D points is propagated along the iso-depth contour to recover the complete 3D shape. The spatially varying BRDF is computed once the 3D shape is recovered.


\subsection{Analysis and Ablation}\label{ss:statistical_anl}

\begin{figure}[t]
    \begin{center}
      \includegraphics[width=0.23\textwidth]{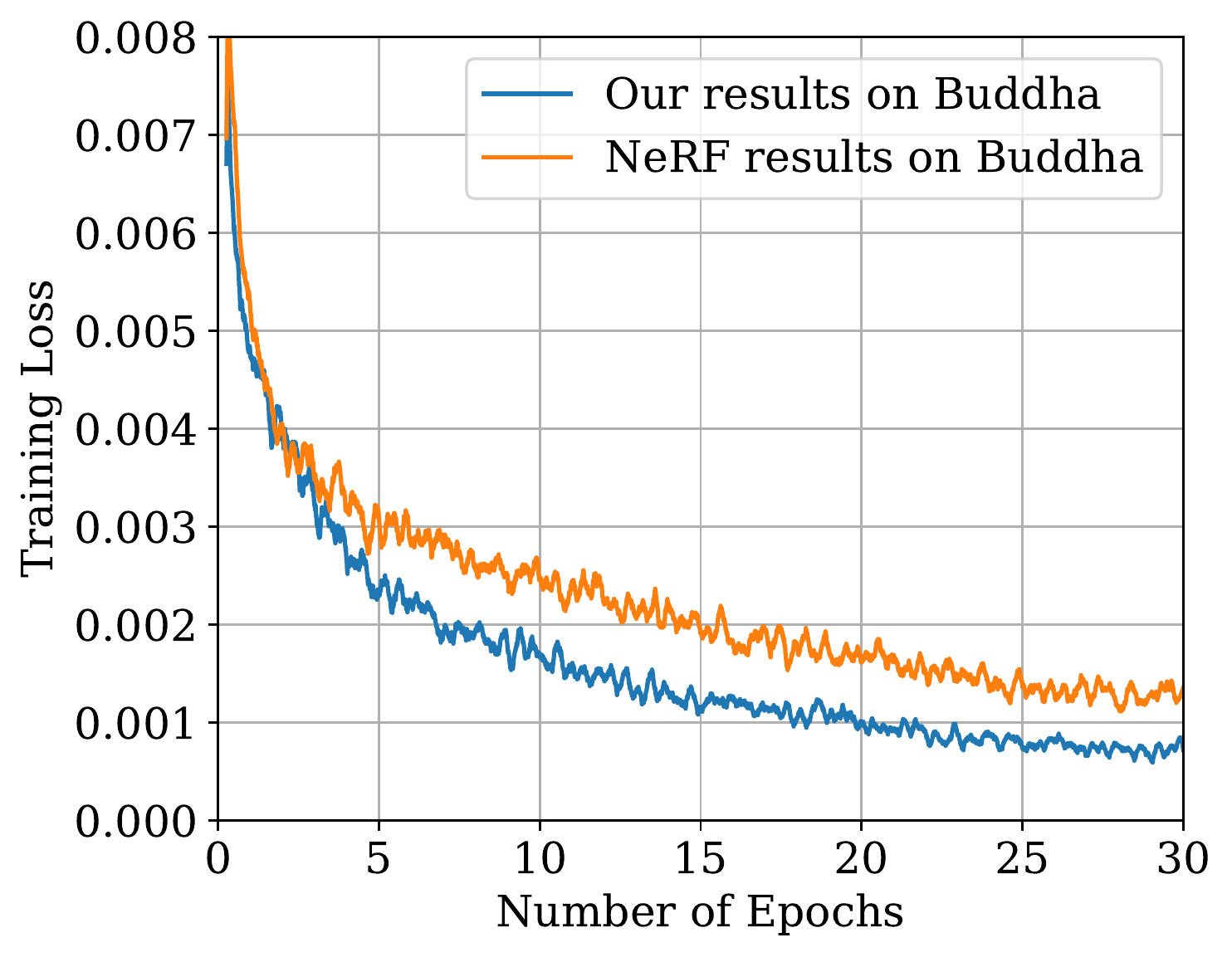}
      \includegraphics[width=0.23\textwidth]{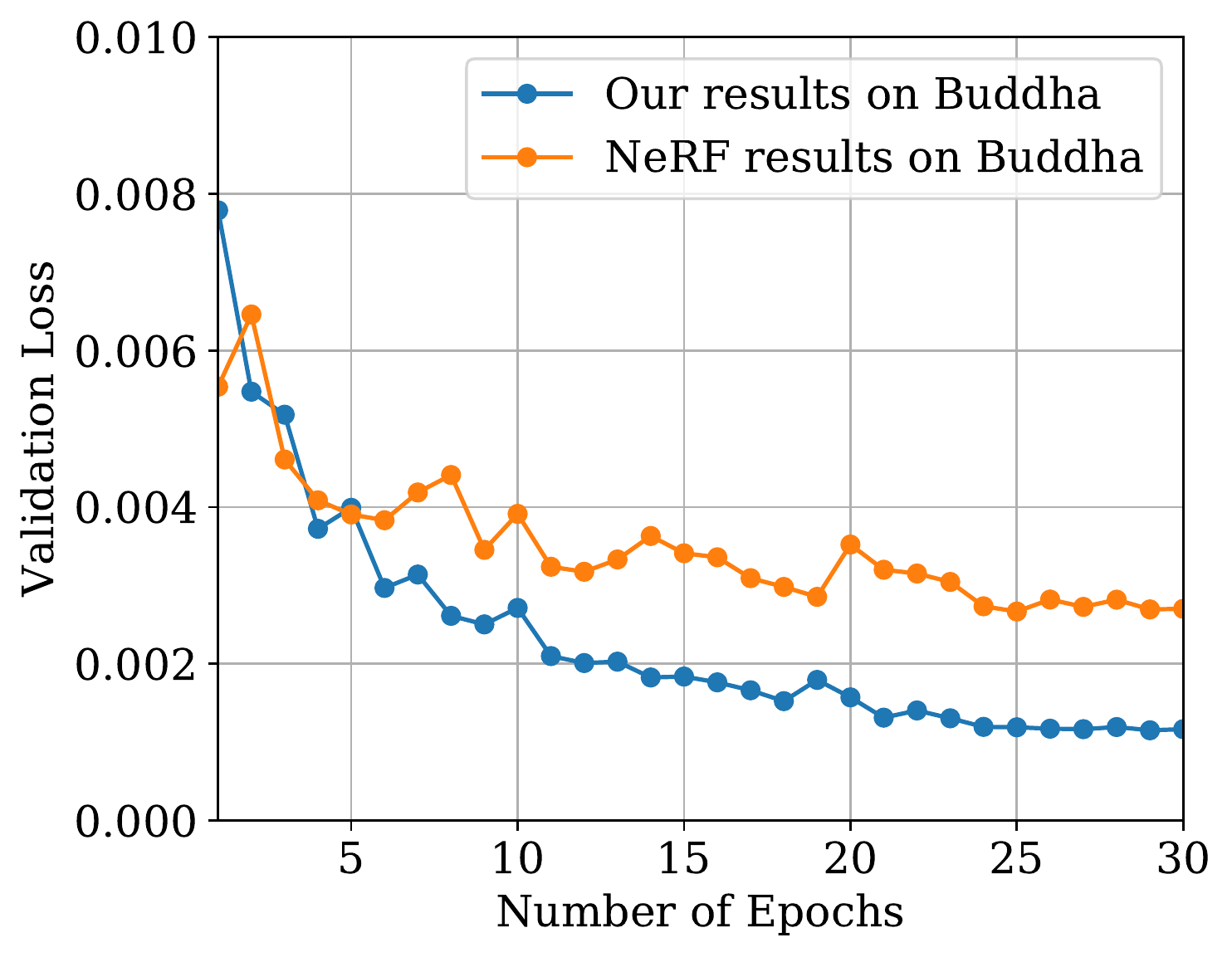}
    \end{center}
    \caption{\small Training and validation loss curve on BUDDHA seq.}\label{fig:PSNR_curves}
\end{figure}

\begin{figure}[t]
    \centering
    \includegraphics[{width=1.0\linewidth}]{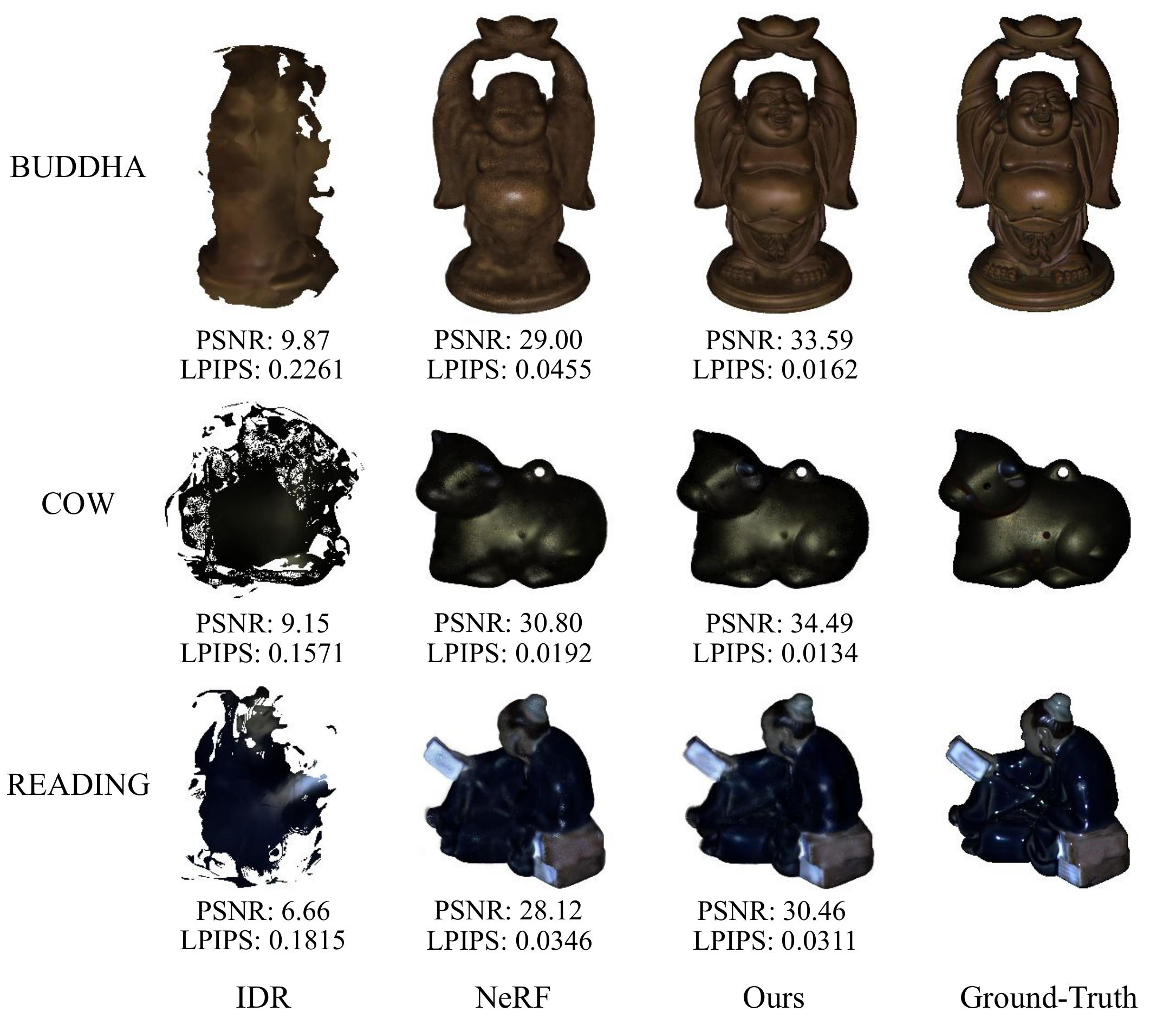}
    \caption{ \small Comparison of volume rendering on DiLiGenT-MV achieved by IDR \cite{yariv2020multiview}, NeRF \cite{mildenhall2020nerf} and our method. Without surface normals, NeRF lacks in details and produces blurry renderings. On the other hand, our method recovers fine surface details and renders accurate images by blending surface normal information in the volume rendering process. PSNR (higher the better), LPIPS (lower the better).
    }
    \label{fig:renderings}
\end{figure}

\noindent \textit{(a) Training and Validation Analysis:}
To demonstrate the effect of surface normal information on image rendering, we compare our method with NeRF.  Fig.\ref{fig:PSNR_curves} provides training and validation curves for both methods on BUDDHA. As expected, our method provides much higher image rendering quality during the learning process. Additionally, Table \ref{tab:psnr_lpips} provides PSNR and LPIPS \cite{zhang2018unreasonable} scores for IDR, NeRF and our method. These results show that our rendering quality is much better than other view-synthesis approaches.

\noindent \textit{(b) Effect of Volume Sampling:} We test the importance of the sampling strategy by performing an ablation study on the BEAR dataset.
In the first experiment, we reduced the number of coarse samples $N_c$ to 16. This increased our reconstruction error from 0.66 to 0.84, indicating that an accurate coarse geometry is needed for better performance.
As a second experiment, we increased $N_c$ to 256, but we did not use fine sampling. Here, the reconstruction error remained at 0.66, showing that more samples must get similar reconstruction quality.


\noindent \textit{(c) Effects of multiple light sources on NeRF:}
Since our method uses multiple light sources for normal estimation using deep PS, one could think about supplying multiple light source images to NeRF. So, rather than choosing the single best light source and apply NeRF to multi-view images, we test here what happens if we provide multiple light source images to it.
The motivation of this experiment is to study the generalization of NeRF to \textit{(i)} different light sources, however, for each experiment the light is consistent throughout the image sequence.
\textit{(ii)} what if the source is changing over the image sequence of the dataset. 
For this experiment, we sampled 20 lights out of 96 and experimented on the DiLiGenT-MV dataset. The minimal condition for PS to work is having more than three images \cite{woodham1980photometric} under Lambertian assumptions, and therefore, 20 light source is good enough. Table \ref{tab:20_lights} shows the empirical results for setup \textit{(i)}. For more results, see supp. material.

\begin{table}[t]
 \scriptsize
	\centering
		\begin{tabular}{c|c|c|c|c|c}
			\hline
			\rowcolor[gray]{0.85}
			Dataset & BEAR & BUDDHA & COW  & POT2 & READING \\ \hline
			NeRF \cite{mildenhall2020nerf} & 0.80  & 1.07 & \textbf{0.78} & 0.81   & 1.18  \\ \hline
			Ours & \textbf{0.70} & \textbf{1.06} &  0.79 & \textbf{0.73} & \textbf{0.98} \\ \hline
		\end{tabular}
		\caption{\small Quantitative 3D reconstruction accuracy against NeRF \cite{mildenhall2020nerf}. We tested both approaches with 20 different light configurations. We provide average scores using Chamfer-L1 metric.}\label{tab:20_lights}
\end{table}

\formattedparagraph{Limitations and Further Study.}\label{ss:limitations}
Our method combines two independent research fields that practice precise 3D reconstruction of an object from images. We demonstrated that the proposed method provides favorable results against the competing methods. However. we make a few assumptions, such as calibrated MVPS setup and solid objects, which limit our approach to broader adoption. Further, we do not explicitly model inter-reflections. Consequently, it would be interesting to extend our work to the uncalibrated settings. Also, exploring joint modeling of PS normal and surface normal from the implicit surface representation is left as a future extension. To improve the convergence time and modeling of uncertainty, we are in process of assessing \cite{lindell2020autoint, barron2021mip, yu2021plenoctrees, garbin2021fastnerf, deng2021depth}.



\section{Conclusion}\label{sec:conclusion}
We introduced a straightforward method that takes an opportunistic approach to solve the multi-view photometric stereo problem. We conclude that by exploiting the photometric stereo image formation model and the recent continuous volume rendering for multi-view image synthesis, we can reconstruct the 3D geometry of the object with high accuracy. Further, our formulation inherently utilizes the notion of the density gradient by leveraging the photometric stereo response and hence bypasses the explicit modeling of 3D occupancy information.
Of course, using prior from structure-from-motion and its variants could be beneficial, as it could provide surface occupancy directly, but it would make the framework more complex.
We demonstrated that introducing knowledge about the density gradient to the neural radiance field representation provides encouraging improvements in MVPS. We assessed the suitability of our work via extensive experiments on the DiLiGenT-MV benchmark dataset. We believe that our work can open up a new direction for study in multi-view photometric stereo for further developments.

\smallskip
\formattedparagraph{Acknowledgement.} {{This work was funded by Focused Research Award from Google (CVL, ETH 2019-HE-318, 2019-HE-323).}}

{\small
\bibliographystyle{ieee_fullname}
\bibliography{wacv_arxiv}
}

\appendix
\begingroup
\let\clearpage\relax 
\onecolumn 

\section*{\centering{\Large Neural Radiance Fields Approach to Deep Multi-View Photometric Stereo \newline [Supplementary Material]}}

\begin{abstract}



Here, we extend the experimental section from the main paper. Our supplementary document is organized as follows: First, we present the visual comparison of the rendered images with our method and other view synthesis methods. We also show the advantage of using surface normals via training and validation curves.  The following sections provide further analysis by comprehensive experiments. Concretely, we demonstrate the effect of change in the light source direction between the frames on NeRF \cite{mildenhall2020nerf}. We also investigate the importance of viewing directions and sampling strategy on our method. Later, we provide some additional implementation details related to our method. \textbf{Finally, we reemphasize the main motivation of this work in the concluding remarks section}.

\end{abstract}

\section{Comparison with View Synthesis Methods }

In this section, we compare rendering performances of view-synthesis methods.

\formattedparagraph{Visual Comparison.} Fig.\ref{fig:rendering_comparison}  shows the images rendered by IDR \cite{yariv2020multiview}, NeRF \cite{mildenhall2020nerf} and our method. All the methods use the DiLiGenT-MV images captured with the same light source. We observed that the IDR method fails to capture the geometry and appearance information with this setting. The NeRF method performs much better than IDR; however, the rendered images are often blurry and lack surface details. We observed that our method performs significantly better than both methods, and it can generate important details of the object. For example, our approach renders the nose of the BEAR and the eyes of the BUDDHA very accurately. These details are not apparent with NeRF due to the missing surface normal information in rendering.

\begin{figure*}[t]
    \centering
    \includegraphics[{width=1.0\linewidth}]{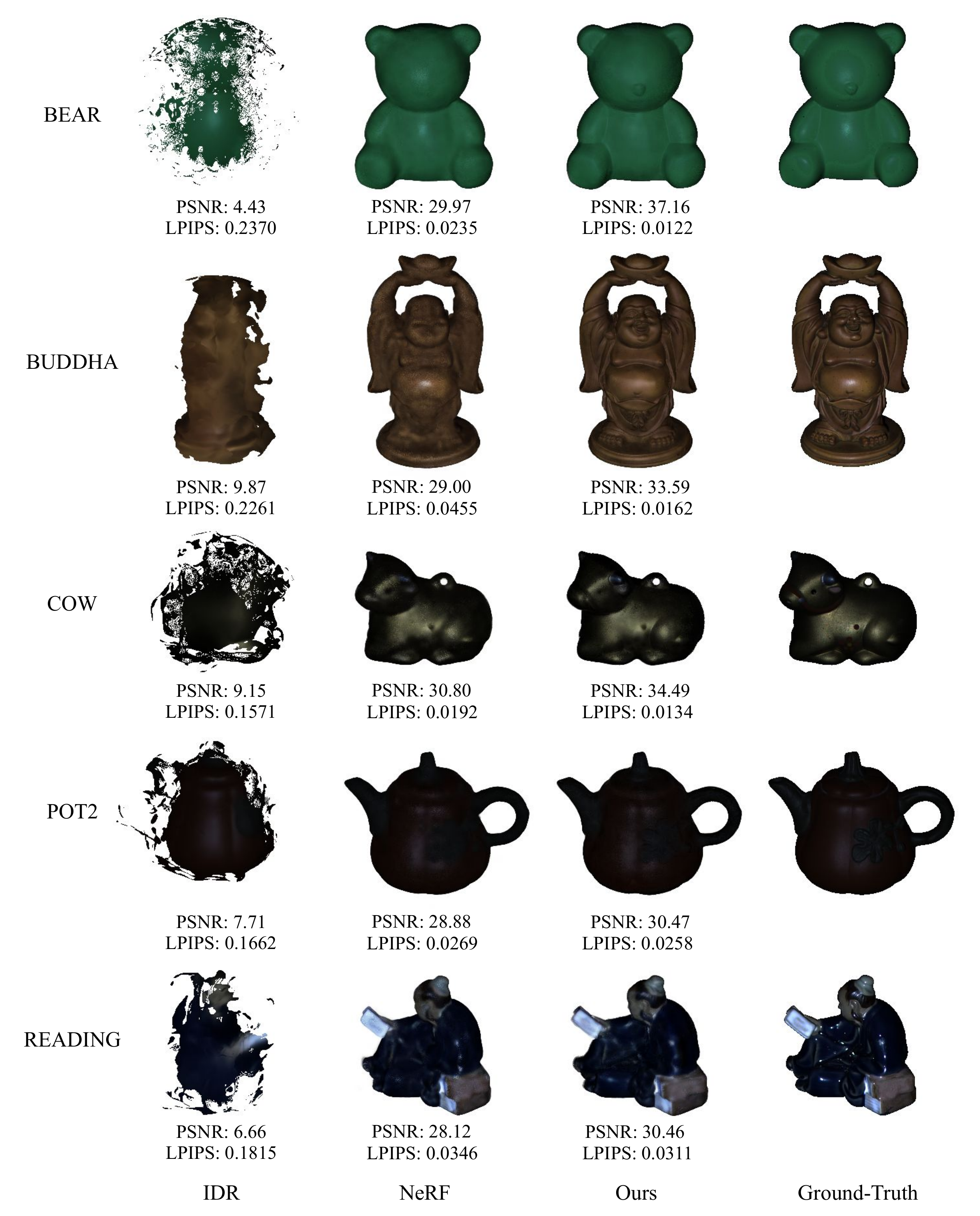}
    \caption{ \small Visual comparison on DiLiGenT-MV renderings achieved by IDR \cite{yariv2020multiview}, NeRF \cite{mildenhall2020nerf} and our method. Without surface normals, NeRF lacks in details and produces blurry renderings. On the other hand, our method is able to recover fine details and render accurate images by blending surface normal information in volume rendering process. 
    We observed that IDR framework cannot recover the geometry and appearance on this benchmark.  }
    \label{fig:rendering_comparison}
\end{figure*}

\formattedparagraph{Training and Validation Analysis.}
Our method combines photometric stereo surface normals in the continuous volume rendering process for better image rendering. The surface normals obtained using the photometric stereo take care of shading in the image formation process. And therefore, it can be observed from the plots presented in Fig.\ref{fig:loss_curves_ours_vs_nerf}, that our method clearly shows better convergence behavior. In addition to the BUDDHA scene presented in the main paper, we analyze the loss curves of the remaining DiLiGenT-MV objects. Fig.\ref{fig:loss_curves_ours_vs_nerf} shows the training and validation curves of our method and NeRF for BEAR, COW, POT2, and READING. Our method has a lower loss value in all of the categories, indicating that the image rendering quality is better.

\begin{figure*}
\centering
\subfigure[\label{fig:bear_loss} BEAR]{\includegraphics[width=0.45\textwidth]{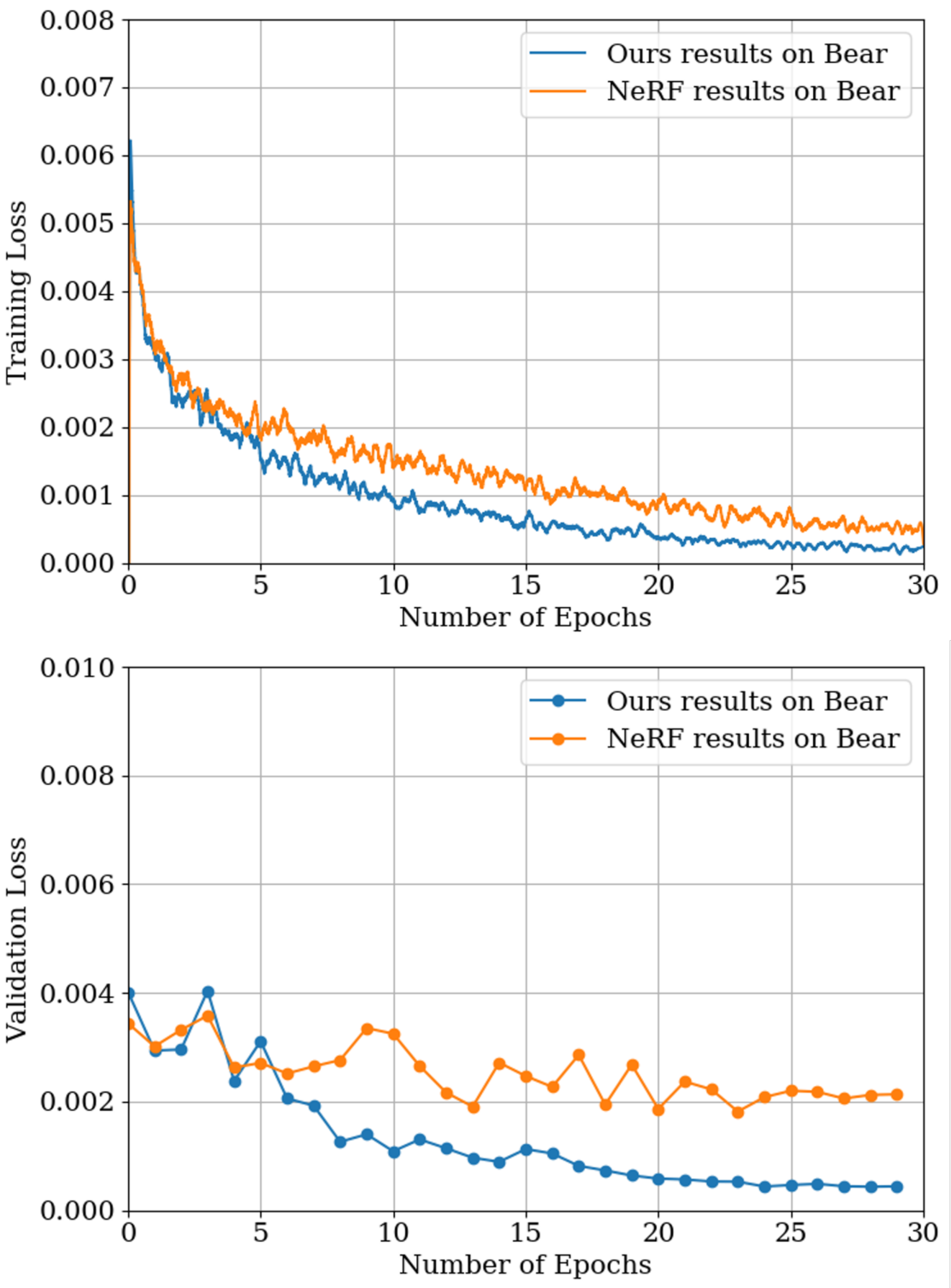}}%
~~~~~~~
\subfigure[\label{fig:cow_loss} COW]{\includegraphics[width=0.45\textwidth]{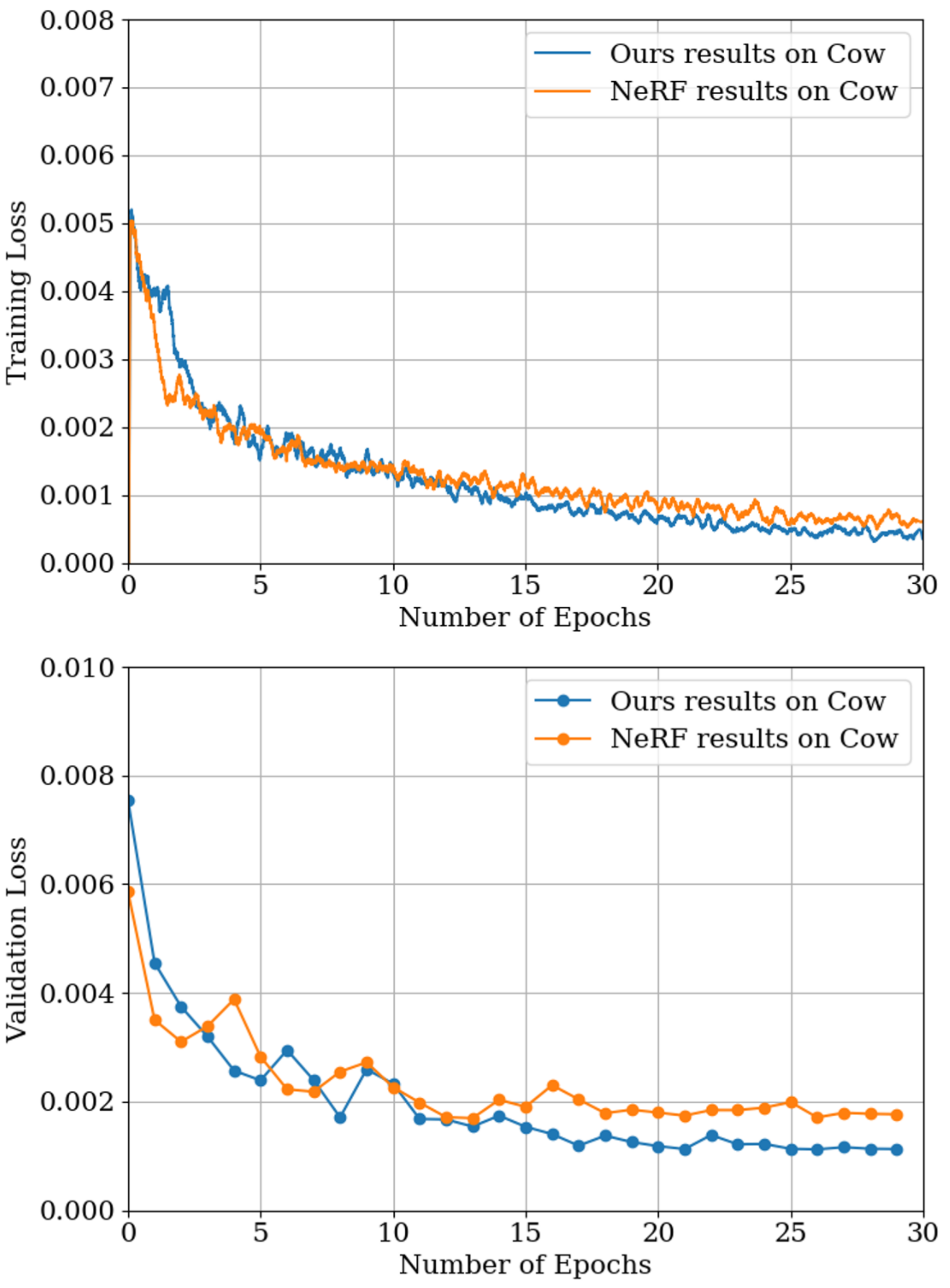}}\\%
\subfigure[\label{fig:pot2_loss}POT2 ]{\includegraphics[width=0.45\textwidth]{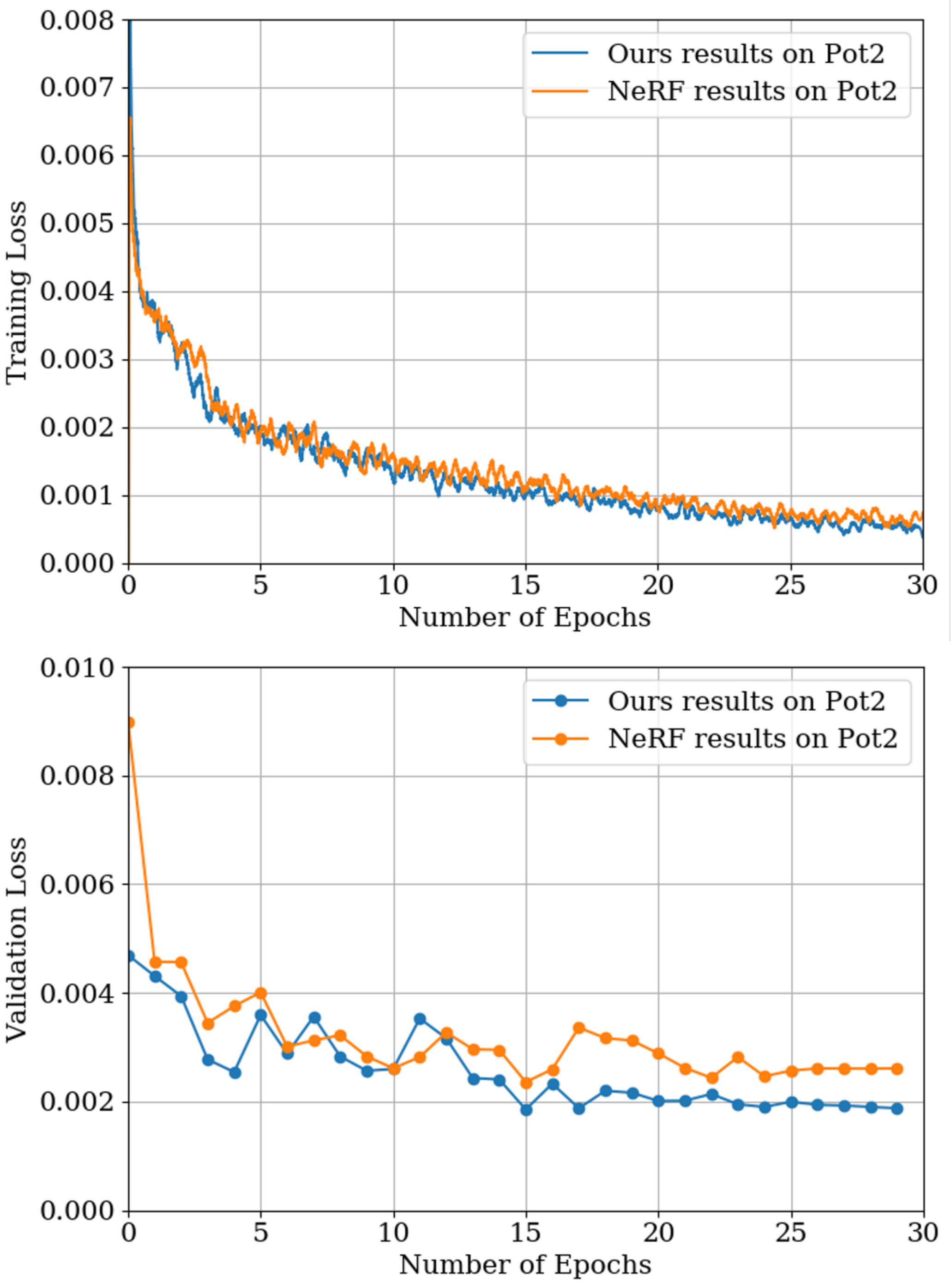}}%
~~~~~~~\subfigure[\label{fig:reading_loss}READING]{\includegraphics[width=0.45\textwidth]{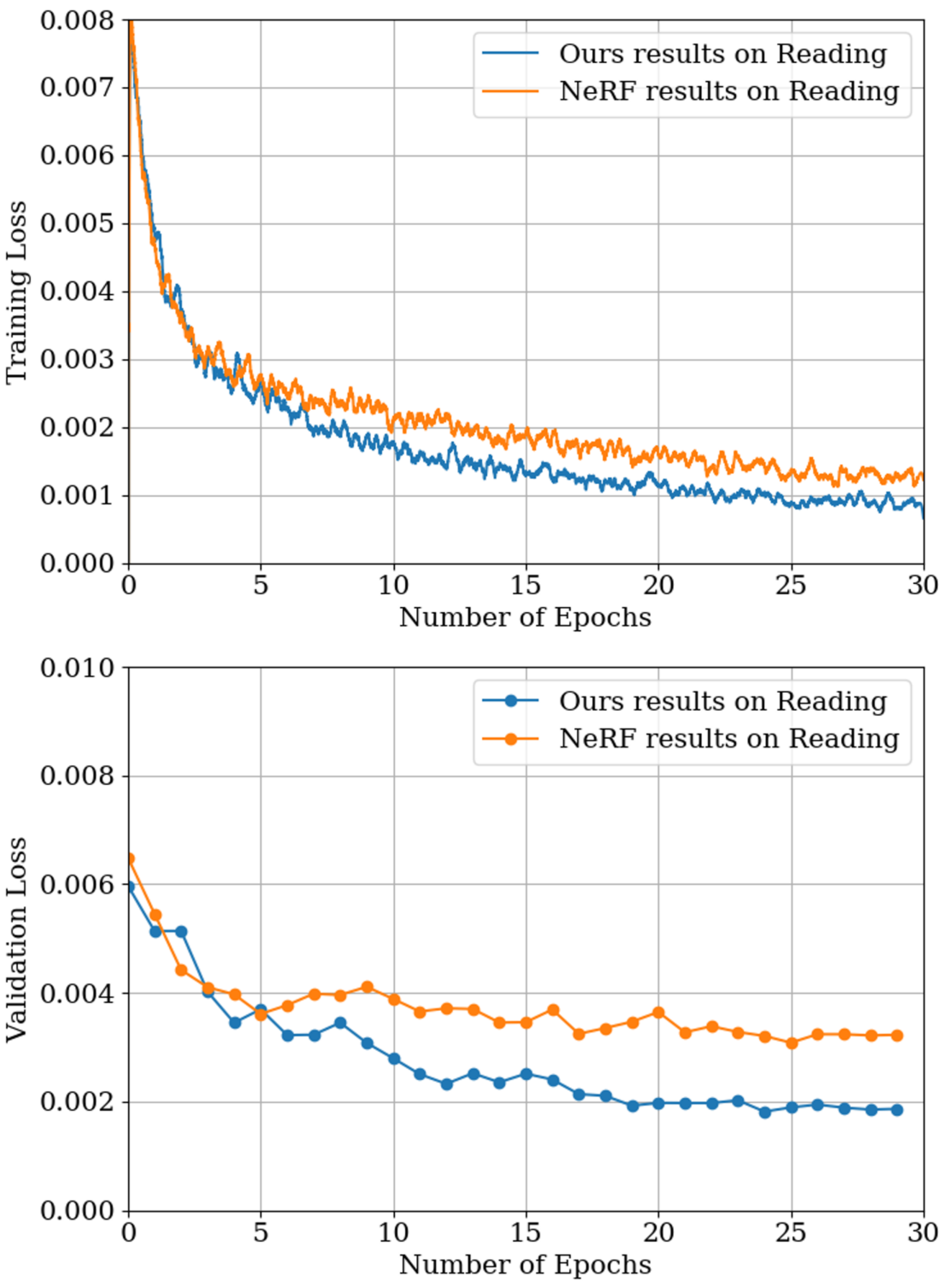}}
\caption{Training and validation curves of BEAR (a), COW(b), POT2(c) and READING(d) using our method and NeRF. Our method consistently shows better convergence behavior with the contribution of the surface normal information.}
\label{fig:loss_curves_ours_vs_nerf}
\end{figure*}

\section{Effect of Multiple Light Sources on NeRF}

We want to study the effect of change in the light source direction over MVPS images. This setting has not been studied before, where the light is different across images that are captured from multiple views.
To simulate this experiment, the subject in the scene should have the appropriate material properties and surface area to show the change of light direction over images. In this paper, to conduct this experiment, we choose the suitable example that has complex surface profile and has significant surface area to really capture the effects of change in the light source direction across images. Hence, we choose COW and BUDDHA as suitable examples from DiLiGenT-MV dataset to simulate this experiment.

NeRF enforces the learned scene representation to be multi-view consistent by learning position-dependent volume density $\sigma$, and it renders images by taking view direction into account. For this reason, having constant lighting on the scene is required, and the core idea of the method is questionable if lighting varies across camera viewpoints.
In this section, we study the behavior of NeRF under multiple light sources. 
For our experiment, we assign a different light source for each camera viewpoint. To be precise, we randomly pick 20 light sources from 96. Table \ref{tab:multiple_light_sources} compares the 3D reconstruction accuracy achieved using images from the same light source and multiple sources. 

\begin{table}[h]
 \scriptsize
	\centering
		\begin{tabular}{c|c|c}
			\hline
			\rowcolor[gray]{0.85}
			Method  & BUDDHA & COW   \\ \hline
			NeRF with single light source &  0.99 &  0.9    \\ \hline
			NeRF with multiple light sources &  1.30 &  1.06 \\ \hline
		\end{tabular}
		\caption{\small Comparison of reconstruction accuracy achieved by NeRF with single light source and multiple light sources. We report Chamfer-L1 distance for the comparison.
		}\label{tab:multiple_light_sources}
\end{table}

        
        

\section{Effect of Viewing Direction}


Here, we want to study the effect of viewing direction on rendered image quality obtained using our method for this experiment. To that end, we remove view direction information $\gamma(\mathbf{d})$ from our MLP. Table \ref{tab:psnr_lpips_no_view_dependence} compares the quality of image rendering with and without view dependence. As expected, the image quality sharply decreases without the view direction. So we conclude that similar to surface normals, view direction is also crucial for the rendering.



\begin{table}[h]
 \scriptsize
	\centering
		\begin{tabular}{c|c|c|c|c|c}
			\hline
			\rowcolor[gray]{0.85}
			Method & BEAR & BUDDHA & COW  & POT2 & READING \\ \hline
			 Ours without view dependence & 31.71 & 29.76 &  30.26 & 30.28 &  29.41 \\ \hline
			Ours with view dependence & \textbf{37.16} &  \textbf{33.59}  & \textbf{34.49}   & \textbf{30.47} & \textbf{30.46}  \\ \hline
		\end{tabular}
		 \caption{\small Quantitative image rendering quality measurement with PSNR metric with and without view dependence (The higher the better). }
		\label{tab:psnr_lpips_no_view_dependence}
\end{table}

\section{Effect of Volume Sampling}


Our method uniformly samples points along the ray between near and far bounds $t_n, t_f$. Increasing the number of these query points enables a denser evaluation of the network. Still, it is computationally not feasible to sample a lot of points uniformly. To make the process more efficient, we use a two-stage hierarchical volume sampling strategy by optimizing coarse and fine networks simultaneously. For that, we first consider the coarse network rendering: 

\begin{equation}
    \centering
    \begin{aligned}
    \Tilde{C}_{c}(\mathbf{r}) = \sum_{i=1}^{N_c}  {\mathbf{w}}_i (\mathbf{x}_i) \mathbf{c}_i (\mathbf{x}_i, \mathbf{n}_i^{ps}, \mathbf{d}) , \text{where} ~\mathbf{w}_i (\mathbf{x}_i) = T_i\Big(1-\exp\big(\sigma(\mathbf{x}_i)\delta_i\big)\Big)
    \end{aligned}\label{eq:coarse_color}
\end{equation}

We calculate weights $\hat{{\mathbf{w}}}_i(\mathbf{x}_i) = {{\mathbf{w}}_i}(\mathbf{x}_i) / \sum_{j=1}^{N_c} {{\mathbf{w}}_j}(\mathbf{x}_j)$ to have a probability density function on the ray. Then, we sample fine points from this distribution using inverse transform sampling. For the coarse network, we sample $N_c = 64$ points uniformly. For the fine network, we sample $N_f = 128$ points by taking the coarse network weights into account.



To show the effectiveness of our two-stage sampling strategy, we simulate an experiment. To that end, we remove fine sampling from our approach and evaluate the performance by using only the uniformly sampled points. Table \ref{tab:sampling_strategy} reports the 3D reconstruction accuracy using 64 and 256 coarse samples only, as well as the two-stage approach of using both coarse and fine sampling. The results suggest that choosing $N_c$ and $N_f$ introduces a better trade-off between computation time and accuracy.

\begin{table}[h]
 \scriptsize
	\centering
		\begin{tabular}{c|c|c|c|c|c}
			\hline
			\rowcolor[gray]{0.85}
			Volume Sampling & BEAR & BUDDHA & COW  & POT2 & READING \\ \hline
			$N_c = 64$, $N_f = 0$ &0.85 &1.45 & 0.86 &0.63 & 1.38\\ \hline
			$N_c = 256$, $N_f = 0$ & 0.68 &0.92 & 1.01&0.64 & 1.64\\ \hline
			$N_c = 64$, $N_f = 128$ & 0.66 &  1.00 &  0.71 & 0.63 & 0.82\\ \hline
		\end{tabular}
		\caption{\small
		 Reconstruction accuracy achieved with different number of points. We provide scores using Chamfer-L1 distance metric.
		}\label{tab:sampling_strategy}
\end{table}

\section{Additional Implementation Details}
\formattedparagraph{Deep Photometric Stereo Network:}
The deep photometric stereo network is trained on the synthetic CyclesPS dataset \cite{ikehata2018cnn}. CyclesPS has 15 different shapes, and for each shape, three sets of images are rendered with varying material properties. For training, the effect of different light sources on each pixel is represented using an observation map. The data is further augmented by applying ten different rotations.
Concretely, the same rotation is applied to the observation maps and corresponding ground-truth normals. The idea is that rotation of surface normals, and light directions around the view direction of photometric stereo setup do not alter the image value for the isotropic surfaces. For more information, we refer the readers to CNN-PS work \cite{ikehata2018cnn}.

The observation maps are rotated ten times by uniformly picking angles in the range $[0, 360]$ during testing. We run inference on each observation map separately and aggregate the results by simply applying inverse rotations. We then average and normalize these vectors to get the per-pixel normal estimate. This strategy improves the accuracy of the normal estimations.


\formattedparagraph{MLP Optimization for MVPS:} Our MLP is implemented using PyTorch \cite{paszke2017automatic}. While comparing our method against other standalone MVS methods,
 we picked the images coming from the same light source in the DiLiGenT-MV dataset. We observed that the fourth light source provides a consistent surface profile image throughout the image sequence. And therefore, we use it for evaluating all MVS methods. Our method and NeRF also require a common threshold value for getting the 3D. For this reason, we extracted meshes using density thresholds of 1, 5, 10, 20, 50, and 100 using NeRF method. We observed that choosing the density threshold of 10 results in the best performance for NeRF, and therefore we applied the same threshold to our method during mesh extraction.


\section{Concluding Remarks}


Firstly, we want to indicate that MVPS is not an ordinary 3D data acquisition setup that can be realized with common commodity cameras. It requires sophisticated hardware, and special care must be taken to calibrate cameras and lights. Only then, it becomes possible to acquire 3D and render scenes accurately.  Secondly, we want to emphasize again that MVPS is generally solved using a sequence of involved steps. Hence, the main motivation of this work is to utilize the modern approach for the classical MVPS problem and explore how far we can go with it (with a framework that is as simple as possible). This work shows that we can get closer to state-of-the-art multi-stage MVPS methods with a much simpler framework by leveraging the continuous volumetric rendering approach. All in all, this work provides a new way to solve MVPS, and maybe working on such ideas can help us come up with a better and a simpler way to recover 3D from MVPS images.

\endgroup

\end{document}